\documentclass[runningheads]{llncs}
\usepackage{epsfig}
\usepackage{graphicx}
\usepackage{amsmath}
\usepackage{amssymb}
\usepackage{caption}
\usepackage[misc,geometry]{ifsym}
\usepackage{booktabs}
\usepackage{multirow}
\usepackage{siunitx}
\usepackage{bm}
\usepackage{subcaption}
\usepackage{mathtools}
\usepackage{caption} 
\usepackage{color}
\usepackage{xcolor}
\definecolor{OliveGreen}{rgb}{0,0.6,0}
\definecolor{SoftRed}{rgb}{1,0.2,0.2}
\usepackage[pagebackref]{hyperref}   
\hypersetup{
     colorlinks=true,
     linkcolor=blue,
     filecolor=blue,
     citecolor = OliveGreen,      
     urlcolor=cyan,
     }

\newcommand{\R}{\mathbb{R}}

\begin{document}
\title{Palmira: A Deep Deformable Network for Instance Segmentation of Dense and Uneven Layouts in Handwritten Manuscripts}
\titlerunning{Palmira}
%
\author{S P Sharan\orcidID{0000-0002-6298-6464} \and
Sowmya Aitha \orcidID{0000-0003-2266-9947} \and
Amandeep Kumar \orcidID{0000-0002-9292-9778} \and
Abhishek Trivedi \orcidID{0000-0002-6763-4716} \and
Aaron Augustine \orcidID{0000-0003-3022-9714} \and
Ravi Kiran Sarvadevabhatla (\Letter)\orcidID{0000-0003-4134-1154}
}
\authorrunning{Sharan et al.}
%
\institute{Centre for Visual Information Technology\\International Institute of Information Technology, Hyderabad -- 500032, INDIA \\
\url{https://ihdia.iiit.ac.in/Palmira} \\
\email{ravi.kiran@iiit.ac.in}}
\maketitle              
\begin{abstract}
Handwritten documents are often characterized by dense and uneven layout. Despite advances, standard deep network based approaches for semantic layout segmentation are not robust to complex deformations seen across semantic regions. This phenomenon is especially pronounced for the low-resource Indic palm-leaf manuscript domain. To address the issue, we first introduce Indiscapes2, a new large-scale diverse dataset of Indic manuscripts with semantic layout annotations. Indiscapes2 contains documents from four different historical collections and is $150\%$ larger than its predecessor, Indiscapes. We also propose a novel deep network \textsc{Palmira} for robust, deformation-aware instance segmentation of regions in handwritten manuscripts. We also report Hausdorff distance and its variants as a boundary-aware performance measure. Our experiments demonstrate that \textsc{Palmira} provides robust layouts, outperforms strong baseline approaches and ablative variants. We also include qualitative results on Arabic, South-East Asian and Hebrew historical manuscripts to showcase the  generalization capability of \textsc{Palmira}. 
\end{abstract}

\section{Introduction}

Across cultures of the world, ancient handwritten manuscripts are a precious form of heritage and often serve as definitive sources of provenance for a wide variety of historical events and cultural markers. Consequently, a number of research efforts have been initiated worldwide~\cite{clausner2019icdar2019,prusty2019indiscapes,monnier2020docExtractor,Kesiman_2020} to parse image-based versions of such manuscripts in terms of their structural (layout) and linguistic (text) aspects. In many cases, accurate layout prediction greatly facilitates downstream processes such as OCR~\cite{ma2020joint}. Therefore, we focus on the problem of obtaining high-quality layout predictions in handwritten manuscripts.

Among the varieties of historical manuscripts, many from the Indian subcontinent and South-east Asia are written on palm-leaves. These manuscripts pose significant and unique challenges for the problem of layout prediction. The digital versions often reflect multiple degradations of the original. Also, a large variety exists in terms of script language, aspect ratios and density of text and non-text region categories. The Indiscapes Indic manuscript dataset and the deep-learning based layout parsing model by  Prusty et. al.~\cite{prusty2019indiscapes} represent a significant first step towards addressing the concerns mentioned above in a scalable manner. Although Indiscapes is the largest available annotated dataset of its kind, it contains a rather small set of documents sourced from two collections. The deficiency is also reflected in the layout prediction quality of the associated deep learning model. 

To address these shortcomings, we introduce Indiscapes2 dataset as an expanded version of Indiscapes (Sec.~\ref{sec:indiscapes2}). Indiscapes2 is $\textbf{150}\%$ larger compared to its predecessor and contains two additional annotated collections which greatly increase qualitative diversity (see Fig.~\ref{fig:indiscapes2}, Table~\ref{table:document-counts}). In addition, we introduce a novel deep learning based layout parsing architecture called Palm leaf Manuscript Region Annotator or \textsc{Palmira} in short (Sec.~\ref{sec:palmira}). Through our experiments, we show that \textsc{Palmira} outperforms the previous approach and strong baselines, qualitatively and quantitatively (Sec.~\ref{sec:results}). Additionally, we demonstrate the general nature of our approach on out-of-dataset historical manuscripts. Intersection-over-Union (IoU) and mean Average Precision (AP) are popular measures for scoring the quality of layout predictions~\cite{He2017MaskR}. Complementing these area-centric measures, we report the boundary-centric Hausdorff distance and its variants as part of our evaluation approach (Sec.~\ref{sec:results}). 

The source code, pretrained models and associated material are available at this link: \url{https://ihdia.iiit.ac.in/Palmira}.

\section{Related Work}

Layout analysis is an actively studied problem in the document image analysis community~\cite{clausner2019icdar2019,lee2019page,barman2020combining}. For an overview of approaches employed for historical and modern document layout analysis, refer to the work of Prusty et. al.~\cite{prusty2019indiscapes} and Liang et. al.~\cite{liang2018efficient}. In recent times, large-scale datasets such as PubLayNet~\cite{zhong2019publaynet} and DocBank~\cite{li2020docbank} have been introduced for document image layout analysis. These datasets focus on layout segmentation of modern language printed magazines and scientific documents.  

Among recent approaches for historical documents, Ma et. al.~\cite{ma2020joint} introduce a unified deep learning approach for layout parsing and recognition of Chinese characters in a historical document collection. Alaasam et. al.~\cite{alaasam2019layout} use a Siamese Network to segment challenging historical Arabic manuscripts into main text, side text and background. Alberti et. al.~\cite{alberti2019labeling} use a multi-stage hybrid approach for segmenting text lines in medieval manuscripts. Monnier et. al.~\cite{monnier2020docExtractor} introduce docExtractor, an off-the-shelf pipeline for historical document element extraction from 9 different kinds of documents utilizing a modified U-Net~\cite{ronneberger2015u}. dhSegment ~\cite{oliveira2018dhsegment} is a similar work utilizing a modified U-Net for document segmentation of medieval era documents. Unlike our instance segmentation formulation (i.e. a pixel can simultaneously have two distinct region labels), existing works (except dhSegment) adopt the classical segmentation formulation (i.e. each pixel has a single region label). Also, our end-to-end approach produces page and region boundaries in a single stage end-to-end manner without any postprocessing.

Approaches for palm-leaf manuscript analysis have been mostly confined to South-East Asian scripts~\cite{valy2018character,paulus2018improved} and tend to focus on the problem of segmented character recognition~\cite{9287584,puarungroj2020using,Kesiman_2020}. The first large-scale dataset for palm leaf manuscripts was introduced by Prusty et. al.~\cite{prusty2019indiscapes}, which we build upon to create an even larger and more diverse dataset.

Among deep-learning based works in document understanding, using deformable convolutions~\cite{dai2017deformable} to enable better processing of distorted layouts is a popular choice. However, existing works have focused only on tabular regions~\cite{Siddiqui2018DeCNTDD,agarwal2020cdec}. We adopt deformable convolutions, but for the more general problem of multi-category region segmentation.

\begin{table}[!t]
\caption{Document collection level, region level statistics of Indiscapes2 dataset.} 
    \begin{subtable}[t]{.35\linewidth}
    \captionof{table}{Collection level stats.}
    \resizebox{\linewidth}{!}
{
    \centering
    \begin{tabular}{c|ccc|c|c}
 \toprule 
  & Train  &  Validation & Test  & Total & Indiscapes (old)\\
\midrule   
\textsc{PIH}         & $285$        & $70$        & $94$           & $\mathbf{449}$ & $193$   \\\textsc{Bhoomi}      & $408$        & $72$         & $96$           & $\mathbf{576}$       & $315$   \\
 \textsc{ASR}      & $36$        & $11$         & $14$           & $\mathbf{61}$        & $--$  \\
 \textsc{jain}      & $95$        & $40$         & $54$           & $\mathbf{189}$         & $--$ \\
    \midrule   
    Total      & $824$       & $193$         & $258$           & $\mathbf{1275}$          & $508$  \\
 \bottomrule
 \end{tabular}
 }
\label{table:document-counts} 
    \end{subtable}
    \begin{subtable}[t]{.65\linewidth}
    \captionof{table}{Region count statistics.}
    \resizebox{\linewidth}{!}
{
    \centering
    \begin{tabular}{c|c|c|c|c|c|c|c|c|c}
\midrule 
 & Character   & Character  &  Hole     & Hole       & Page & Library  & Decorator/&  Physical & Boundary \\
 & Line Segment & Component & (Virtual) & (Physical) & Boundary & Marker & Picture & Degradation & Line \\
 & (CLS)  & (CC)  & (Hv) & (Hp) & (PB) & (LM) & (D/P) & (PD) & (BL) \\
 \midrule
\textsc{PIH}  & $5105$ & $1079$ & $-$ & $9$ & $610$ & $52$ & $153$ & $90$ & $724$ \\
\textsc{Bhoomi}      & $5359$ & $524$ & $8$ & $737$ & $547$ & $254$ & $8$ & $2535$& $80$\\
\textsc{ASR}      & $673$ & $59$ & $-$ & $-$ & $52$ & $41$ & $-$ & $81$& $83$\\
\textsc{Jain}   & $1857$ & $313$ & $93$ & $38$ & $166$ & $7$ & $-$ & $166$& $292$\\
\midrule
Combined      & $12994$ & $1975$ & $101$ & $784$ & $1375$ & $354$ & $161$ & $2872$& $1179$ \\
 \bottomrule
 \end{tabular}
 }
\label{tab:dataset-region-stats} 
    \end{subtable}
\end{table}

\section{Indiscapes2}
\label{sec:indiscapes2}

We first provide a brief overview of Indiscapes dataset introduced by Prusty et. al.~\cite{prusty2019indiscapes}. This dataset contains $508$ layout annotated manuscripts across two collections - Penn-in-Hand (from University of Pennsylvania’s Rare Book Library) and Bhoomi (from libraries and Oriental Research Institutes across India). The images span multiple scripts, exhibit diversity in language and text line density, contain multiple manuscripts stacked in a single image and often contain  non textual elements (pictures and binding holes). 

Although Indiscapes was the first large-scale Indic manuscript dataset, it is rather small by typical dataset standards. To address this limitation and enable advanced layout segmentation deep networks, we build upon Indiscapes to create Indiscapes2. For annotation, we deployed an instance of HInDoLA~\cite{trivedi2019hindola} - a multi-feature annotation and analytics platform for historical manuscript layout processing. The fully automatic layout segmentation approach from Prusty et. al.~\cite{prusty2019indiscapes} is available as an annotation feature in HInDoLA. The annotators utilize the same to obtain an initial estimate and edit the resulting output, thus minimizing the large quantum of labour involved in pure manual annotation. HInDoLA also provides a visualization interface for examining the accuracy of annotations and tagging documents for correction.

In the new dataset, we introduce additional annotated documents from the Penn-in-Hand and Bhoomi book collections mentioned previously. Above this, we also add annotated manuscripts from two new collections - ASR and Jain. The ASR documents are from a private collection and contain $61$ manuscripts written in Telugu language. They contain $18-20$ densely spaced text lines per document (see Fig.~\ref{fig:indiscapes2}). The Jain collection contains $189$ images. These documents contain $16-17$ lines  per page and include early paper-based documents in addition to palm-leaf manuscripts.  

\begin{figure*}[!t]
    \centering
    \includegraphics[width=\textwidth]{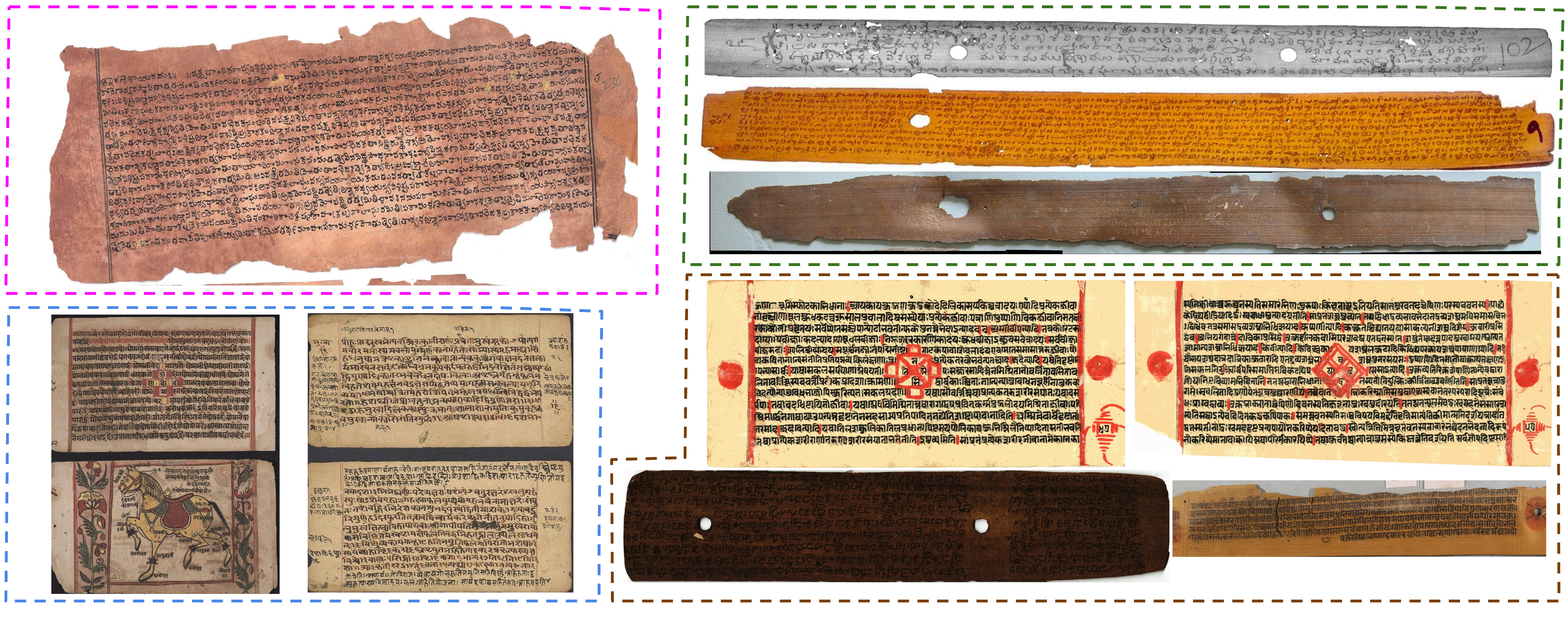}
    \caption{Representative manuscript images from Indiscapes2 - from newly added ASR collection (top left, pink dotted line), Penn-in-Hand (bottom left, blue dotted line), Bhoomi (green dotted line), newly added Jain (brown dotted line). Note the diversity across collections in terms of document quality, region density, aspect ratio and non-textual elements (pictures).}
    \label{fig:indiscapes2}
\end{figure*}

Altogether, Indiscapes2 comprises of $1275$ documents - a $150\%$ increase over the earlier Indiscapes dataset. Refer to Tables~\ref{table:document-counts},\ref{tab:dataset-region-stats} for additional statistics related to the datasets and Fig.~\ref{fig:indiscapes2} for representative images. Overall, Indiscapes2 enables a greater coverage across the spectrum of historical manuscripts - qualitatively and quantitatively.

\section{Our Layout Parsing Network (\textsc{Palmira})}
\label{sec:palmira}

\begin{figure*}[!t]
    \centering
    \includegraphics[width=\textwidth]{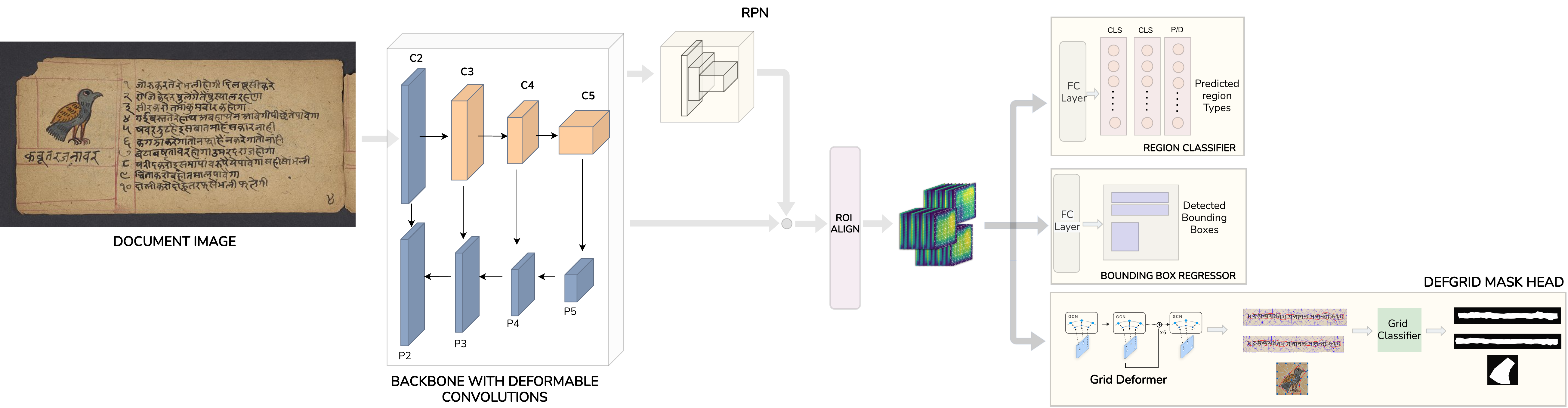}
    \caption{A diagram illustrating \textsc{Palmira}'s architecture (Sec.~\ref{sec:palmira}). The orange blocks in the backbone are deformable convolutions (Sec.~\ref{sec:mod1-defconv}). Refer to Fig.~\ref{fig:defgrid}, Sec.~\ref{sec:mod2-defgrid} for additional details on Deformable Grid Mask Head which outputs region instance masks.}
    \label{fig:indiscapes2}
\end{figure*}

In their work, Prusty et. al.~\cite{prusty2019indiscapes} utilize a modified Mask-RCNN~\cite{He2017MaskR} framework for the problem of localizing document region instances. Although the introduced framework is reasonably effective, the fundamental convolution operation throughout the Mask-RCNN deep network pipeline operates on a fixed, rigid spatial grid. This rigidity of receptive fields tends to act as a bottleneck in obtaining precise boundary estimates of manuscript images containing highly deformed regions. To address this shortcoming, we modify two crucial stages of the Mask R-CNN pipeline in a novel fashion to obtain our proposed architecture (see Fig.~\ref{fig:indiscapes2}). To begin with, we  briefly summarize the Mask R-CNN approach adopted by Prusty et. al. We shall refer to this model as the Vanilla Mask-RCNN model. Subsequently, we shall describe our novel modifications to the pipeline. 
    
\subsection{Vanilla Mask-RCNN}

Mask R-CNN~\cite{He2017MaskR} is a three stage deep network for object instance segmentation. The three stages are often referred to as Backbone, Region Proposal Network (RPN) and Multi-task Branch Networks. One of the Branch Networks, referred to as the Mask Head, outputs individual object instances. The pipeline components of Mask-RCNN are modified to better suit the manuscript image domain by Prusty et al~\cite{prusty2019indiscapes}. 
Specifically, the ResNet-50 used in Backbone is initialized from a Mask R-CNN trained on the MS-COCO dataset. Within the RPN module, the anchor aspect ratios of 1:1,1:3,1:10  were chosen keeping the peculiar aspect ratios of manuscript images in mind and the number of proposals from RPN were reduced to $512$. The various thresholds involved in other stages (objectness, NMS) were also modified suitably. Some unique modifications were included as well -- the weightage for loss associated with the Mask head was set to twice of that for the other losses and focal-loss~\cite{lin2017focal} was used for robust labelling. 
    
We use the modified pipeline described above as the starting point and incorporate two novel modifications to Mask-RCNN. We describe these modifications in the sections that follow.
    
\subsection{Modification-1: Deformable Convolutions in Backbone}
\label{sec:mod1-defconv}
    
Before examining the more general setting, let us temporarily consider 2D input feature maps $\bm{x}$. Denote the 2D filter operating on this feature map as $\bm{w}$ and the convolution grid operating on the feature map as $\mathcal{R}$. As an example, for a $3 \times 3$ filter, we have:
    
\begin{equation}
            \mathcal{R} =
            \begin{Bmatrix}
            (-1,-1) & (-1,0) & (-1,1)\\
            (0,-1) & (0,0) & (0,1)\\
            (1,-1) & (1,0) & (1,1)\\
            \end{Bmatrix}
\end{equation}
    
\noindent Let the output feature map resulting from the convolution be $\bm{y}$. For each pixel location $\bm{p}_0$, we have:
    
\begin{equation}
        \label{eq:regularconv}
            \bm{y}(\bm{p}_0) = \sum_{\bm{p}_n \in \mathcal{R}} \bm{w}(\bm{p}_n) \cdot \bm{x}(\bm{p}_0 + \bm{p}_n)
\end{equation}
        
\noindent where $n$ indexes the spatial grid locations associated with $\mathcal{R}$. The default convolution operation in Mask R-CNN operates via a fixed 2D spatial integer grid as described above. However, this setup does not enable the grid to deform based on the input feature map, reducing the ability to better model the high inter/intra-region deformations and the features they induce. 
    
As an alternative, Deformable Convolutions~\cite{dai2017deformable} provide a way to determine suitable local 2D offsets for the default spatial sampling locations (see Fig.~\ref{fig:defconv}). Importantly, these offsets $\{\Delta \bm{p}_n ; n=1,2\ldots \}$ are adaptively computed as a function of the input features for each reference location $\bm{p}_0$. Equation \ref{eq:regularconv} becomes:
        
\begin{equation}
            \bm{y}(\bm{p}_0) = \sum_{\bm{p}_n \in \mathcal{R}} \bm{w}(\bm{p}_n) \cdot \bm{x}(\bm{p}_0 + \bm{p}_n + \Delta \bm{p}_n)
        \label{eq:dconv}
\end{equation}        
        
\noindent Since the offsets $\Delta \bm{p}_n$ may be fractional, the sampled values for these locations are generated using bilinear interpolation. This also preserves the differentiability of the filters because the offset gradients are learnt via backpropagation through the bilinear transform. Due to the adaptive sampling of spatial locations, broader and generalized receptive fields are induced in the network. Note that the overall optimization involves jointly learning both the regular filter weights and weights for a small additional set of filters which operate on input to generate the offsets for input feature locations (Fig.~\ref{fig:defconv}). 
        
\subsection{Modification-2: Deforming the spatial grid in Mask Head}
\label{sec:mod2-defgrid}
    
The `Mask Head' in Vanilla Mask-RCNN takes aligned feature maps for each plausible region instance as input and outputs a binary mask corresponding to the predicted document region. In this process, the output is obtained relative to a $28 \times 28$ regular spatial grid representing the entire document image. The output is upsampled to the original document image dimensions to obtain the final region mask. As with the convolution operation discussed in the previous section, performing upsampling relative to a uniform (integer) grid leads to poorly estimated spatial boundaries for document regions, especially for our challenging manuscript scenario. 
    
\begin{figure*}[t!]
    \centering
    \begin{subfigure}[t]{0.45\textwidth}
        \centering
        \includegraphics[width=\textwidth]{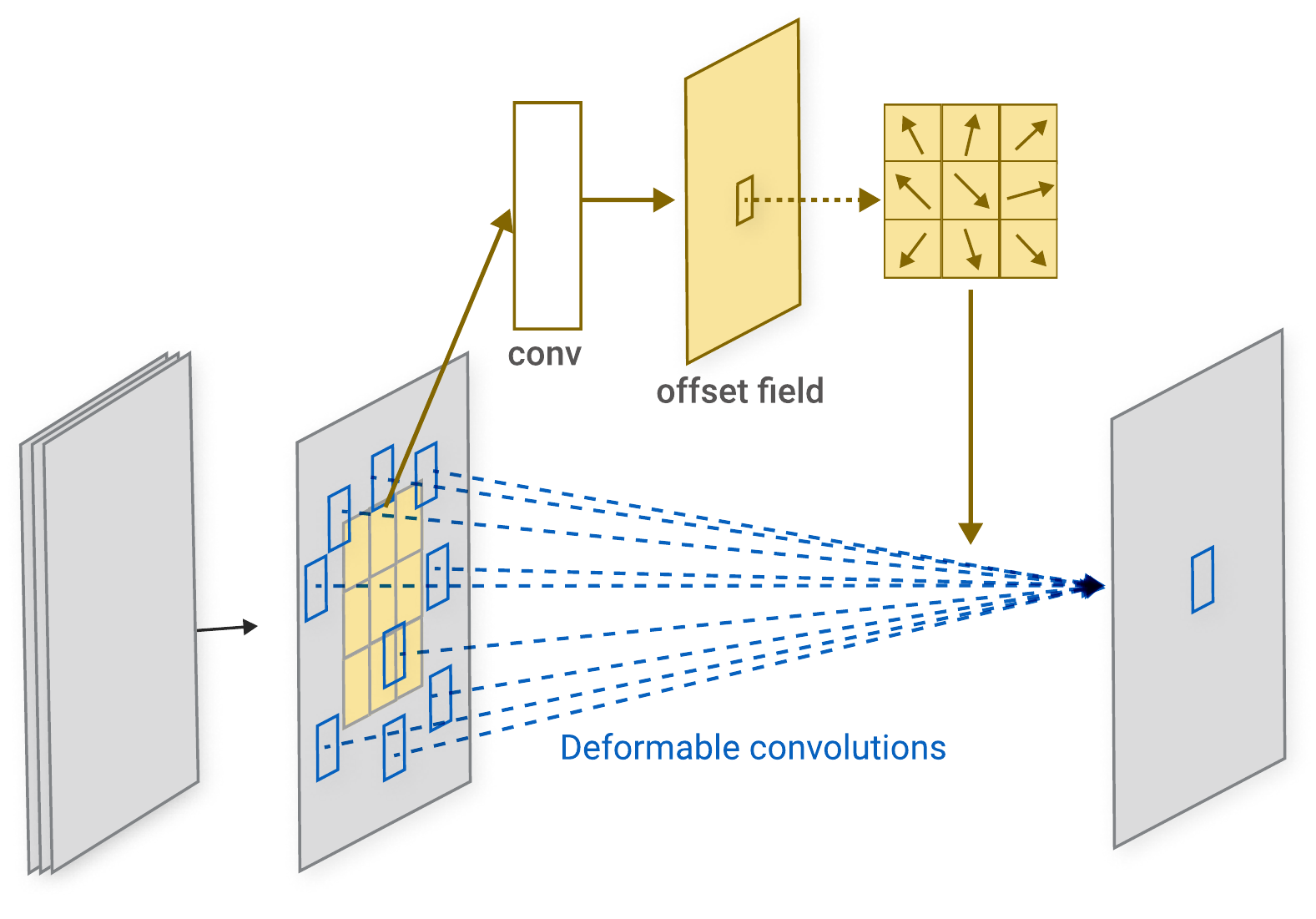}
        \caption{Deformable convolution(Sec.~\ref{sec:mod1-defconv}).}
        \label{fig:defconv}
    \end{subfigure}%
    ~
    \begin{subfigure}[t]{0.55\textwidth}
        \centering
        \includegraphics[width=\textwidth]{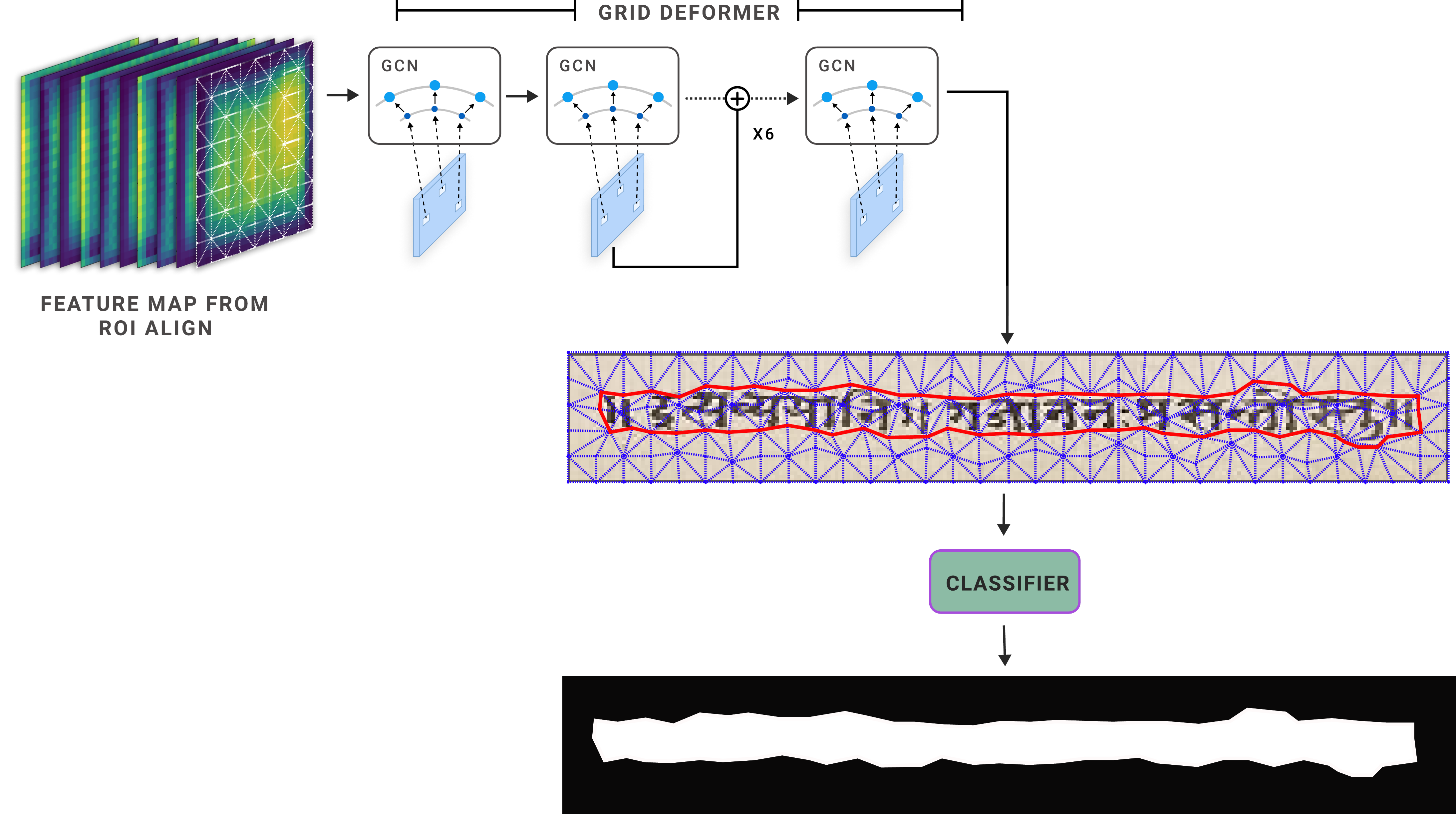}
        \caption{Deformable Grid Mask Head (Sec.~\ref{sec:mod2-defgrid}).}
        \label{fig:defgrid}
    \end{subfigure}
    \caption{Our novel modifications to the Vanilla Mask-RCNN framework (Sec.~\ref{sec:palmira}).}
\end{figure*}

Similar in spirit to deformable convolutions, we adopt an approach wherein the output region boundary is obtained relative to a deformed grid~\cite{gao2020beyond} (see Fig.~\ref{fig:defgrid}). Let $F \in \R^{256\times14\times14}$ be feature map being fed as input to the Mask Head. Denote each of the integer grid vertices that tile the $14\times14$ spatial dimension as $v_i=\left[ x_i,y_i \right]^T$. Each grid vertex is maximally connected to its $8$-nearest neighbors to obtain a grid with triangle edges (see `Feature Map from ROI Align' in Fig.~\ref{fig:defgrid}). The Deformable Grid Mask Head network is optimized to predict the offsets of the grid vertices such that a subset of edges incident on the vertices form a closed contour which aligns with the region boundary. To effectively model the chain-graph structure of the region boundary, the Mask Head utilizes six cascaded Residual Graph Convolutional Network blocks for prediction of offsets. The final layer predicts binary labels relative to the deformed grid structure formed by the offset vertices (i.e. $v_i+[\Delta x_i,\Delta y_i]$). The resulting deformed polygon mask is upsampled to input image dimensions via bilinear interpolation to finally obtain the output region mask. 
    
\subsection{Implementation Details}

\noindent \textit{Architecture:} The Backbone in \textsc{Palmira} consists of a ResNet-50 initialized from a Mask R-CNN network trained on the MS-COCO dataset. Deformable convolutions (Sec.~\ref{sec:mod1-defconv}) are introduced as a drop-in replacement for the deeper layers C3-C5 of the Feature Pyramid Network present in the Backbone (see Fig.~\ref{fig:defconv}). Empirically, we found this choice to provide better results compared to using deformable layers throughout the Backbone. We use $0.5,1,2$ as aspect ratios with anchor sizes of $32,64,128,256,512$ within the Region Proposal Network. While the Region Classifier and Bounding Box heads are the same as one in Vanilla Mask-RCNN (Sec.~\ref{sec:palmira}), the conventional Mask Head is replaced with the Deformable Grid Mask Head as described in Sec.~\ref{sec:mod2-defgrid}. 

\noindent \textit{Optimization:} All input images are resized such that the smallest side is $800$ pixels. 
The mini-batch size is $4$. During training, a horizontal flip augmentation is randomly performed for images in the mini-batch. To address the imbalance in the distribution of region categories (Table~\ref{table:document-counts}), we use repeat factor sampling~\cite{gupta2019lvis} and oversample images containing tail categories. We perform data-parallel optimization distributed across $4$ GeForce RTX 2080 Ti GPUs for a total of $15000$ iterations. A multi-step learning scheduler with warmup phase is used to reach an initial learning rate of $0.02$ after a linear warm-up over $1000$ iterations. The learning rate is decayed by a factor of $10$ at $8000$ and $12000$ iterations. The optimizer used is stochastic gradient descent with gamma $0.1$ and momentum $0.9$. 

Except for the Deformable Grid Mask Head, other output heads (Classifier, Bounding Box) are optimized based on choices made for Vanilla Mask-RCNN~\cite{prusty2019indiscapes}. The optimization within the Deformable Grid Mask Head involves multiple loss functions. Briefly, these loss functions are formulated to (i) minimize the variance of features per grid cell (ii) minimize distortion of input features during differentiable reconstruction (iii) avoid self-intersections by encouraging grid cells to have similar area (iv) encourage neighbor vertices in region localized by a reference central vertex to move in same spatial direction as the central vertex. Please refer to Gao et. al.~\cite{gao2020beyond} for details.

\section{Experimental Setup}
\label{sec:exptsetup}

\subsection{Baselines}

Towards fair evaluation, we consider three strong baseline approaches. 

Boundary Preserving Mask-RCNN~\cite{cheng2020boundary}, proposed as an improvement over Mask-RCNN, focuses on improving the mask boundary along with the task of pixel wise segmentation. To this end, it contains a boundary mask head wherein the mask and boundary are mutually learned by employing feature fusion blocks. 

CondInst~\cite{condInst} is a simple and effective instance segmentation framework which eliminates the need for resizing and RoI-based feature alignment operation present in Mask RCNN. Also, the filters in CondInst Mask Head are dynamically produced and conditioned on the region instances which enables efficient inference. 

In recent years, a number of instance segmentation methods have been proposed as an alternative to Mask-RCNN's proposal-based approach. As a representative example, we use PointRend~\cite{kirillov2020pointrend} - a proposal-free approach. PointRend considers image segmentation as a rendering problem. Instead of predicting labels for each image pixel, PointRend identifies a subset of salient points and extracts features corresponding to these points. It maps these salient point features to the final segmentation label map.

\subsection{Evaluation Setup}

We partition Indiscapes2 dataset into training, validation and test sets (see Table~\ref{table:document-counts}) for training and evaluation of all models, including \textsc{Palmira}. Following standard protocols, we utilize the validation set to determine the best model hyperparameters. For the final evaluation, we merge training and validation set and re-train following the validation-based hyperparameters. A one-time evaluation of the model is performed on the test set. 

\subsection{Evaluation Measures}
    
Intersection-over-Union (IoU) and Average Precision (AP) are two commonly used evaluation measures for instance segmentation.  IoU and AP are area-centric measures which depend on intersection area between ground-truth and predicted masks. To complement these metrics, we also compute boundary-centric measures. Specifically, we use Hausdorff distance (HD)~\cite{hausdorff} as a measure of  boundary precision. For a given region, let us denote the ground-truth annotation polygon by a 2D point set $\mathcal{X}$. Let the prediction counterpart be $\mathcal{Y}$. The Hausdorff Distance between these point sets is given by:
            
            \begin{equation}
                \textup{HD} = d_{H}(\mathcal{X}, \mathcal{Y}) = max \left\{ \adjustlimits \max_{x\in \mathcal{X}} \min_{y\in \mathcal{Y}} d(x,y), \adjustlimits \max_{y\in \mathcal{Y}} \min_{x\in \mathcal{X}} d(x,y) \right\}
            \end{equation}
    
\noindent where $d(x,y)$ denotes the Euclidean distance between points $x \in  \mathcal{X}$, $y \in \mathcal{Y}$. The Hausdorff Distance is sensitive to outliers. To mitigate this effect, the Average Hausdorff Distance is used which measures deviation in terms of a symmetric average across point-pair distances:
    
            \begin{equation}
                \textup{Avg. HD} = d_{AH}(\mathcal{X}, \mathcal{Y}) = \left(\frac{1}{|\mathcal{X}|} \sum_{x\in \mathcal{X}} \min_{y\in \mathcal{Y}} d(x,y) + \frac{1}{|\mathcal{Y}|} \sum_{y\in \mathcal{Y}} \min_{x\in \mathcal{X}} d(x,y)\right)/2            
            \end{equation}

\noindent Note that the two sets may contain unequal number of points ($|\mathcal{X}|,|\mathcal{Y}|$). Additionally, we also compute the $95^{th}$ percentile of Hausdorff Distance ($HD_{95}$) to suppress the effect of outlier distances. 

\begin{table*}[!t]
    
    \captionof{table}{Document-level scores for various performance measures. The baseline models are above the upper separator line while ablative variants are below the line. \textsc{Palmira}'s results are at the table bottom.}
    \resizebox{\textwidth}{!}
    {
        \centering 
        \begin{tabular}{c|c|c|c|c|c|c|c|c}
            \toprule 
            
            Model & Add-On & HD $\downarrow$ & $HD_{95} \downarrow$ & Avg. HD $\downarrow$ & IoU $\uparrow$ & AP $\uparrow$ & $AP_{50} \uparrow$ & $AP_{75} \uparrow$\\
            
            \midrule
            \midrule
            
            PointRend~\cite{kirillov2020pointrend} & - & 252.16 & 211.10 & 56.51 & 69.63 & 41.51 & 66.49 & 43.49 \\
            CondInst~\cite{condInst} & - & 267.73 & 215.33 & 54.92 & 69.49 & 42.39 & 62.18 & 43.03 \\
            Boundary Preserving MaskRCNN~\cite{cheng2020boundary} & - & 261.54 & 218.42 & 54.77 & 69.99 & 42.65 & 68.23 & $\mathbf{44.92}$ \\
            Vanilla MaskRCNN~\cite{prusty2019indiscapes} & - & 270.52 & 228.19 & 56.11 & 68.97 & 41.46 & 68.63 & 34.75 \\
            
            \midrule
            
            Vanilla MaskRCNN & Deformable Convolutions & 229.50 & 202.37 & 51.04 & 65.61 & 41.65 & 65.97 & 44.90 \\
            Vanilla MaskRCNN & Deformable Grid Mask Head  & $\mathbf{179.84}$ & 153.77 & 45.09 & 71.65 & 42.35 & 69.49 & 43.16 \\
            
            \midrule
            
            \textbf{\textsc{Palmira}} : Vanilla MaskRCNN & Deformable Conv., Deformable Grid Mask Head & $184.50$ & $\mathbf{145.27}$ & $\mathbf{38.24}$ &$\mathbf{73.67}$&$\mathbf{42.44}$ & $\mathbf{69.57}$  & 42.93 \\
            
            \bottomrule
        \end{tabular}
    }
    \label{tab:model-results} 
\end{table*}
    
For each region in the test set documents, we compute HD, $HD_{95}$, IoU, AP at IoU thresholds of $50$ ($AP_{50}$) and $75$ ($AP_{75}$). We also compute overall AP by averaging the AP values at various threshold values ranging from $0.5$ to $0.95$ in steps of $0.05$. We evaluate performance at two levels - document-level and region-level. For a reference measure (e.g. HD), we average its values across all regions of a document. The resulting numbers are averaged across all the test documents to obtain document-level score. To obtain region-level scores, the measure values are averaged across all instances which share the same region label. We use document-level scores to compare the overall performance of models. To examine the performance of our model for various region categories, we use region-level scores.

\section{Results}
\label{sec:results}

The performance scores for our approach (\textsc{Palmira}) and baseline models can be viewed in Table~\ref{tab:model-results}. Our approach clearly outperforms the baselines across the reported measures. Note that the improvement is especially apparent for the boundary-centric measures ($HD$, $HD_{95}$, Avg. HD). As an ablation study, we also evaluated variants of \textsc{Palmira} wherein the introduced modifications were removed separately. The corresponding results in Table~\ref{tab:model-results} demonstrate the collective importance of our novel modifications over the Vanilla Mask-RCNN model. 

\begin{table*}[!t]
\captionof{table}{Document-level scores summarized at collection level for various performance measures.}
    \resizebox{\textwidth}{!}
    {
        \centering 
        \begin{tabular}{c|c|c|c|c|c|c|c|c}
            \toprule
            
            Collection name & \# of test images & HD $\downarrow$ & $HD_{95} \downarrow$ & Avg. HD $\downarrow$ & IoU $\uparrow$ & AP $\uparrow$ & $AP_{50} \uparrow$ & $AP_{75} \uparrow$ \\
            
            \midrule
            
            \textsc{PIH} & 94 & 66.23 & 46.51 & 11.16 & 76.78 & 37.57 & 59.68 & 37.63 \\
            \textsc{Bhoomi} & 96 & 220.38 & 175.52 & 46.75 & 69.83 & 30.40 & 50.53 & 29.03 \\
            \textsc{ASR} & 14 & 629.30 & 562.19 & 169.03 & 67.80 & 51.02 & 73.09 & 64.27 \\
            \textsc{Jain} & 54 & 215.14 & 159.88 & 38.91 & 76.59 & 48.25 & 70.15 & 50.34 \\
            
            \midrule
            
            \textsc{Overall} & 258 & 184.50 & 145.27 & 38.24 & 73.67 & 42.44 & 69.57 & 42.93  \\
            \bottomrule
        \end{tabular}
     }
    
    \label{tab:collection-wise-results} 
\end{table*}

\begin{table*}[!t]
    \captionof{table}{\textsc{Palmira}'s overall and region-wise scores for various performance measures. The HD-based measures (smaller the better) are separated from the usual measures (IoU, AP etc.) by a separator line.}
    \resizebox{\textwidth}{!}
    {
        \centering 
        \begin{tabular}{c|c|c|c|c|c|c|c|c|c|c}
            \toprule
            
             & & Character   & Character  &  Hole     & Hole       & Page & Library  & Decorator/&  Physical & Boundary \\
            Metric & \textbf{Overall} & Line Segment & Component & (Virtual) & (Physical) & Boundary & Marker & Picture & Degradation & Line \\
             &  & (CLS)  & (CC)  & (Hv) & (Hp) & (PB) & (LM) & (D/P) & (PD) & (BL) \\
            
            \midrule
            \midrule
            \textsc{${HD}_{95}$} & $\mathbf{171.44}$ & 34.03 & 347.94 & 70.79 & 88.33 & 52.01 & 289.81 & 593.99 & 851.02 & 111.97 \\
            \textsc{Avg HD} & $\mathbf{45.88}$ & 8.43 & 103.98 & 18.80 & 16.82 & 13.19 & 73.23 & 135.46 & 255.86 & 17.95 \\
            
            \midrule
            
            \textsc{IoU (\%)} & $\mathbf{72.21}$ & 78.01 & 54.95 & 74.85 & 77.21 & 92.97 & 67.24 & 50.57 & 27.68 & 61.54 \\
            \textsc{AP} & $\mathbf{42.44}$ & 58.64 & 28.76 & 45.57 & 56.13 & 90.08 & 27.75 & 32.20 & 03.09 & 39.72 \\
            \textsc{$AP_{50}$} & $\mathbf{69.57}$ & 92.73 & 64.55 & 81.20 & 90.53 & 93.99 & 55.18 & 54.23 & 12.47 & 81.24 \\
            \textsc{$AP_{75}$} & $\mathbf{42.93}$ & 92.74 & 64.55 & 81.20 & 90.52 & 93.99 & 55.18 & 54.24 & 12.47 & 81.24 \\
            \bottomrule
        \end{tabular}
    }
   \label{tab:region-results}
\end{table*}

To understand the results at collection level, we summarize the document-level scores of \textsc{Palmira} in Table~\ref{tab:collection-wise-results}. While the results across collections are mostly consistent with overall average, the scores for ASR are suboptimal. This is due to the unsually closely spaced lines and the level of degradation encountered for these documents. It is easy to see that reporting scores in this manner is useful for identifying collections to focus on,for improvement in future.

\begin{figure*}[!t]
    \centering
    \includegraphics[width=\textwidth]{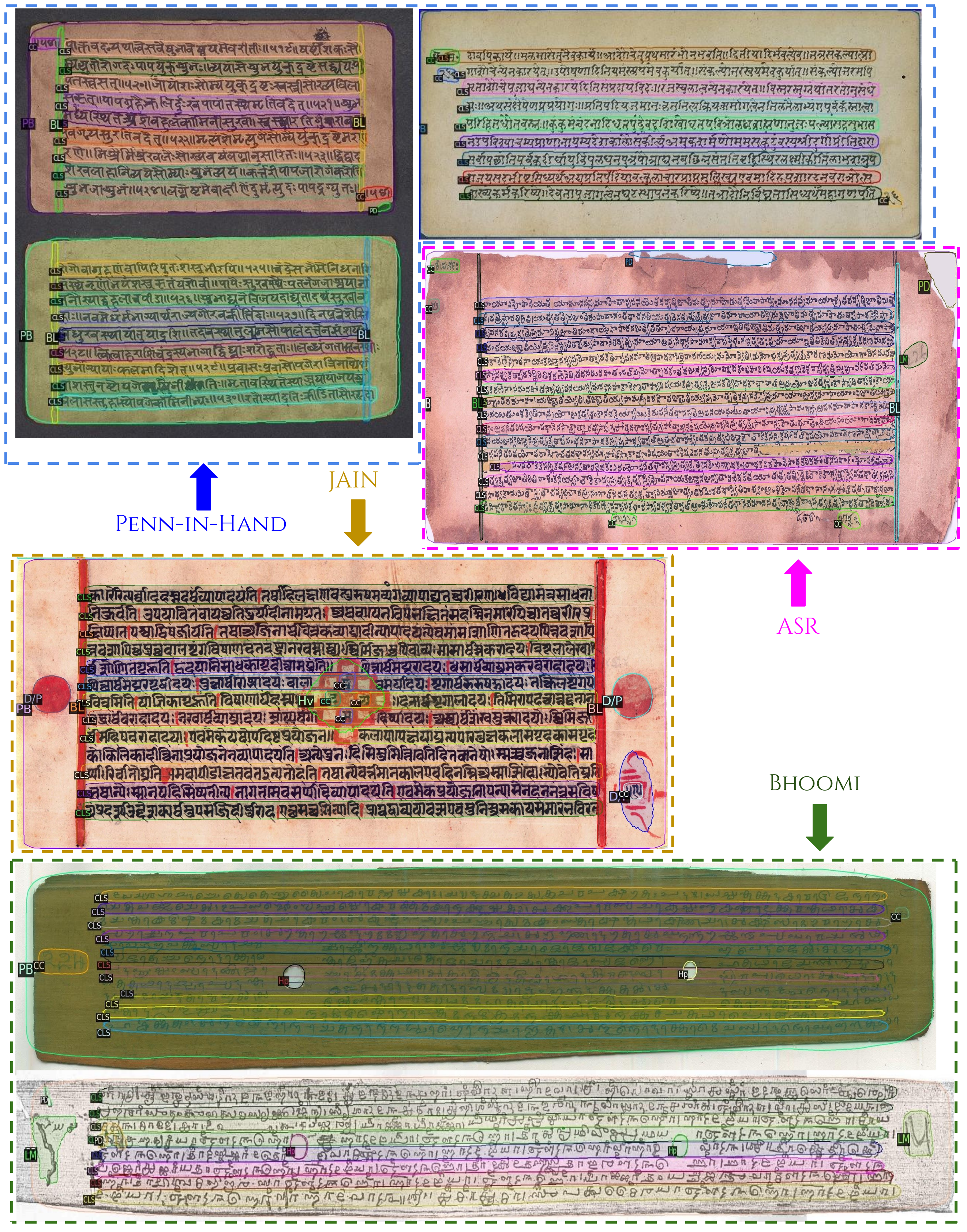}
    \caption{Layout predictions by \textsc{Palmira} on representative test set documents from Indiscapes2 dataset. Note that the colors are used to distinguish region instances. The region category abbreviations are present at corners of the regions.}
    \label{fig:indiscapes2-qualitative}
\end{figure*}

\begin{figure*}[!t]
    \centering
    \includegraphics[width=\textwidth]{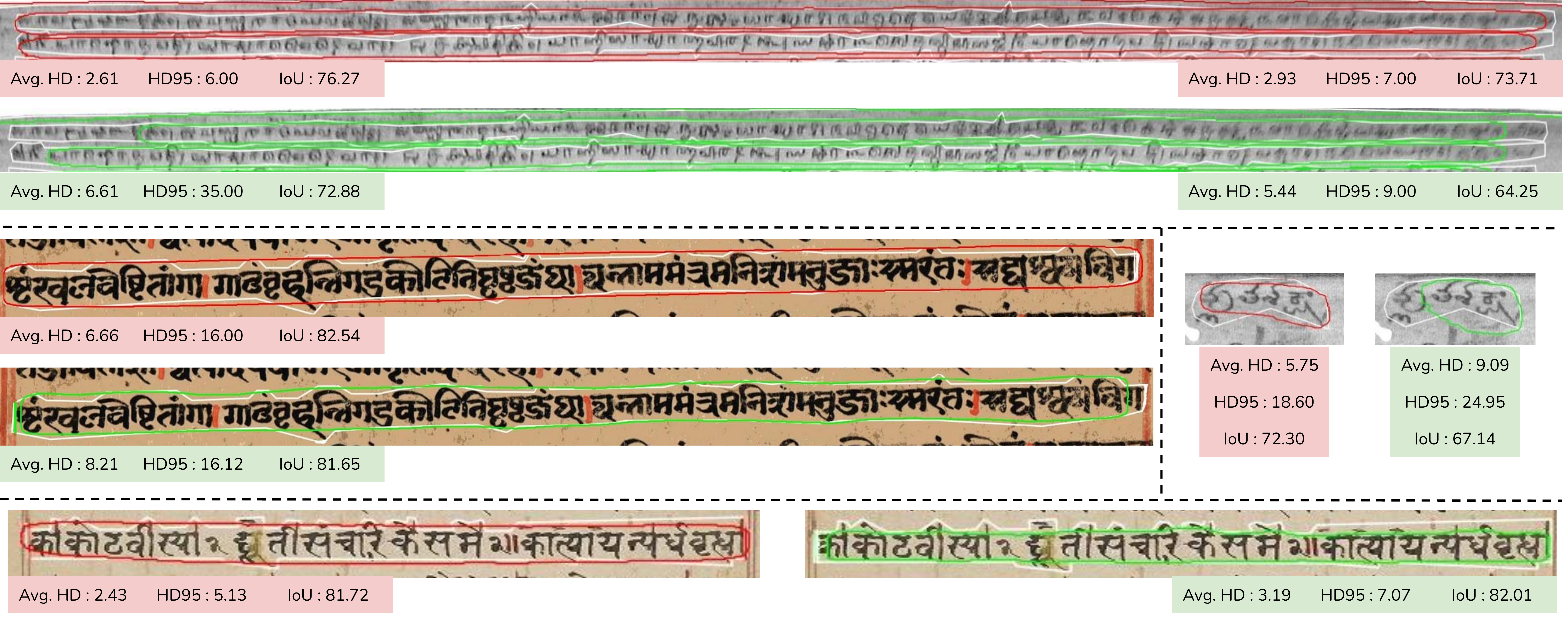}
    \caption{A comparative illustration of region-level performance. \textsc{Palmira}'s predictions are in red. Predictions from the best model among baselines (Boundary-Preserving Mask-RCNN) are in green. Ground-truth boundary is depicted in white.}
    \label{fig:baseline_ours_comparision}
\end{figure*}

We also report the performance measures for \textsc{Palmira}, but now at a per-region level, in Table~\ref{tab:region-results}. In terms of the boundary-centric measures ($HD_{95}$, Avg. HD), the best performance is seen for the most important and dominant region category - Character Line Segment. The seemingly large scores for some categories (`Picture/Decorator', `Physical Degradation') are due to the drastically small number of region instances for these categories. Note that the scores for other categories are reasonably good in terms of boundary-centric measures as well as the regular ones (IoU, AP). 

\begin{figure*}[!t]
    \centering
    \includegraphics[width=\textwidth]{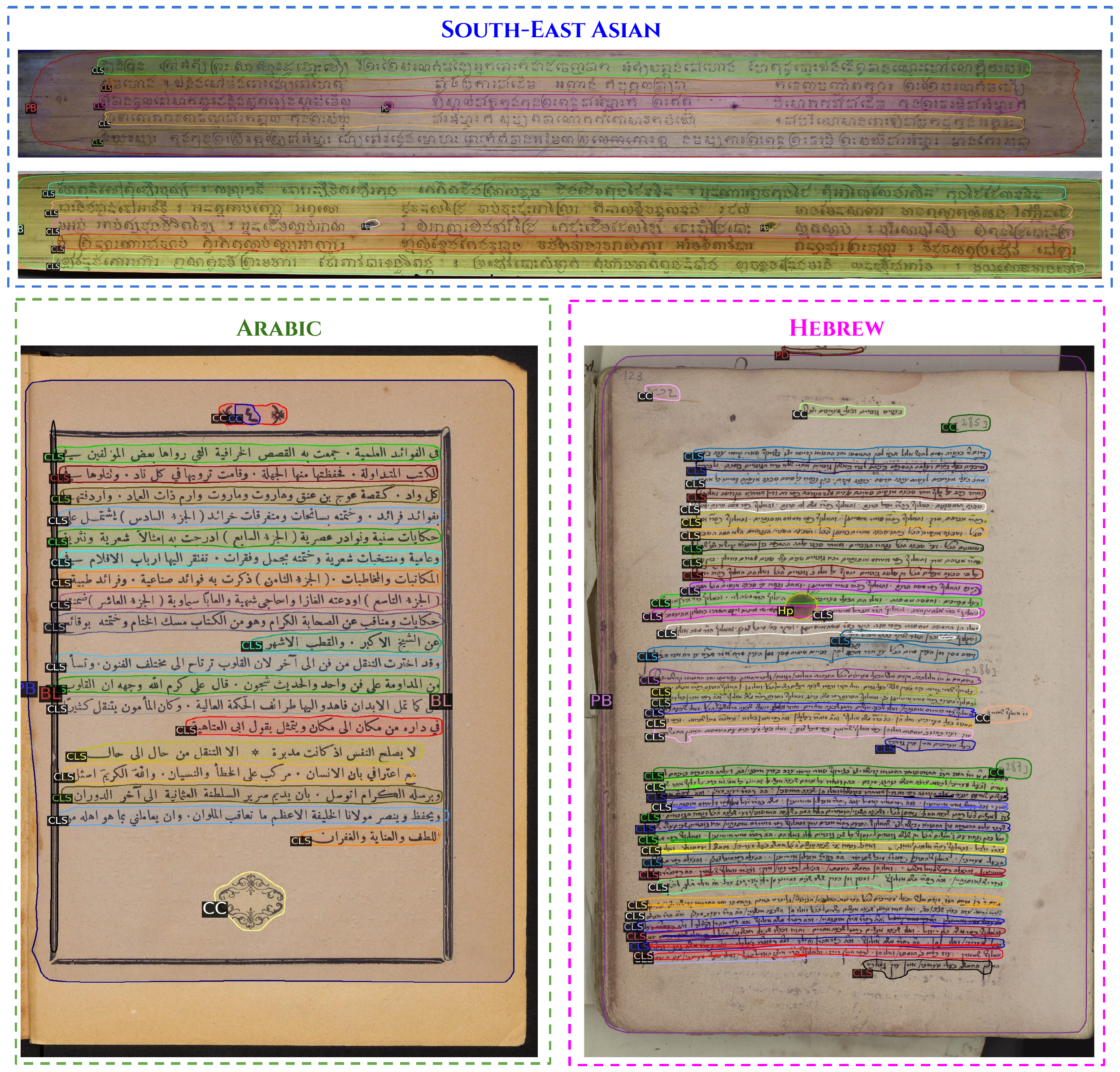}
    \caption{Layout predictions by \textsc{Palmira} on out-of-dataset handwritten manuscripts.}
    \label{fig:indiscapes2-qualitative-others}
\end{figure*}

A qualitative perspective on the results can be obtained from Figure~\ref{fig:indiscapes2-qualitative}. Despite the challenges in the dataset, the results show that \textsc{Palmira} outputs good quality region predictions across a variety of document types. A comparative illustration of region-level performance can be viewed in Figure~\ref{fig:baseline_ours_comparision}. In general, it can be seen that \textsc{Palmira}'s predictions are closer to ground-truth. Figure~\ref{fig:indiscapes2-qualitative-others} shows \textsc{Palmira}'s output for sample South-East Asian, Arabic and Hebrew historical manuscripts. It is important to note that the languages and aspect ratio (portrait) of these documents is starkly different from the typical landscape-like aspect ratio of manuscripts used for training our model. Clearly, the results demonstrate that \textsc{Palmira} readily generalizes to out of dataset manuscripts without requiring additional training.

\section{Conclusion}

There are three major contributions from our work presented in this paper. The \textit{first} contribution is the creation of Indiscapes2, a new diverse and challenging dataset for handwritten manuscript document images which is $150\%$ larger than its predecessor, Indiscapes. The \textit{second} contribution is \textsc{Palmira}, a novel deep network architecture for fully automatic region-level instance segmentation of handwritten documents containing dense and uneven layouts. The \textit{third} contribution is to propose Hausdorff Distance and its variants as a boundary-aware measure for characterizing the performance of document region boundary prediction approaches. Our experiments demonstrate that \textsc{Palmira} generates accurate layouts, outperforms strong baselines and ablative variants. We also demonstrate \textsc{Palmira}'s out-of-dataset generalization ability via predictions on South-East Asian, Arabic and Hebrew manuscripts. Going ahead, we plan to incorporate downstream processing modules (e.g. OCR) for an end-to-end optimization. We also hope our contributions assist in advancing robust layout estimation for handwritten documents from other domains and settings.

\section*{Acknowledgment}
We wish to acknowledge the efforts of all annotators who contributed to the creation of Indiscapes2. 

\bibliographystyle{splncs04}
\bibliography{main}

\end{document}